\documentclass{article}

     \usepackage[final]{neurips_2019}

\usepackage[utf8]{inputenc} %
\usepackage[T1]{fontenc}    %
\usepackage{hyperref}       %
\usepackage{url}            %
\usepackage{booktabs}       %
\usepackage{amsfonts}       %
\usepackage{nicefrac}       %
\usepackage{microtype}      %

\usepackage{longtable}%

\usepackage{booktabs}
\usepackage[load-configurations=version-1]{siunitx} %

\usepackage{times}  %
\usepackage{helvet}  %
\usepackage{courier}  %
\usepackage{url}  %
\usepackage{graphicx}  %
\usepackage{amsmath}
\usepackage{amssymb}
\usepackage{graphicx}
\usepackage{bbm}
\usepackage[ruled]{algorithm2e}
\usepackage{float}

\def\TableDemographics{
\small
\begin{tabular}{lc}
\toprule
Feature & \emph{n}=1,010\\
\midrule
Age & 72 [59, 81]\\
Gender & \\
\quad Female & 466 (0.46)\\
\quad Male & 544 (0.54)\\
Ventricular shunt & 152 (0.15)\\
GCS & 10 [7, 14] \\
Decreased GCS, count & 3,119 \\
Decreased GCS, rate & 1 event / 1.5 days\\
\bottomrule 
\end{tabular}
}

\def\ConceptPseudocode{
\begin{figure}[t]
\begin{minipage}[t]{0.33\textwidth}
\vspace{0pt}
\includegraphics[width=\columnwidth]{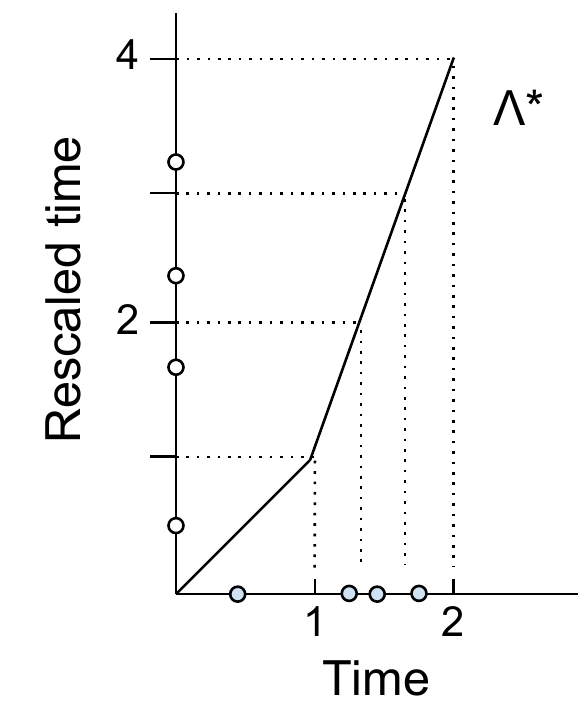}
\end{minipage}%
\hfill
\begin{minipage}[t]{0.62\textwidth}
\vspace{0pt}
\begin{algorithm}[H]
\caption{Harmonic Mean Point Processes} \label{alg_hmpp}
\KwResult{ALL-trained model}
 Temporal network $F: X \mapsto [0,\infty)$ \;
 
 Attention coefficient $\gamma$, stability factor $\epsilon$ \;
 
 \While{training}{
  $\hat{\lambda}(t'_{j,k-1}) = F(X_j)$ piecewise on $[t'_{j,k-1},t'_{j,k})\ \forall k\in K$ \;
  
  Copy then detach $\hat{\lambda}(t'_{j,k-1})\ \forall j, k$ \;
  
  $\text{ALL}_{j,k} = \text{LL}_{j,k} / \big(\hat{\lambda}(t'_{j,k-1})^{\log_{10} \gamma} + \epsilon\big)$\;
  
  ALL.sum().backward()\;
  
  optimizer.step()\;
 }
\end{algorithm}
\end{minipage}%
\caption{A unit of time with $\lambda=1$ has three times less likelihood weight than a unit with $\lambda=3$, evidenced by its proportion of the length on the y-axis (left). Pseudocode for HMPPs (right).} \label{fig:rescale}
\end{figure}
}

\begin{document}

\title{Harmonic Mean Point Processes: 
Proportional Rate Error Minimization %
for Obtundation Prediction}

\author{%
  Yoonjung Kim and Jeremy C. Weiss %
    \\
  Heinz College of Information Systems and Public Policy\\
  Carnegie Mellon University\\
  Pittsburgh, PA 15213 \\
  \texttt{\{yoonjungkim,jeremyweiss\}@cmu.edu} \\
}
\maketitle

\begin{abstract}
In healthcare, the highest risk individuals for morbidity and mortality are rarely those with the greatest modifiable risk.  By contrast, many machine learning formulations implicitly attend to the highest risk individuals. We focus on this problem in point processes, a popular modeling technique for the analysis of the temporal event sequences in electronic health records (EHR) data with applications in risk stratification and risk score systems. We show that optimization of the log-likelihood function also gives disproportionate attention to high risk individuals and leads to poor prediction results for low risk individuals compared to ones at high risk. We characterize the problem and propose an adjusted log-likelihood formulation as a new objective for point processes. 
We demonstrate the benefits of our method in simulations and in EHR data of patients admitted to the critical care unit for intracerebral hemorrhage. %

\end{abstract}

\section{Introduction}
Clinical forecasting is a central task for prognostication in populations at risk for downstream morbidity and mortality.
When followup is incomplete and right-censorship of data occurs, survival analysis and point process models are often preferable to binary classification for long-term prognostication due to their ability to mitigate selection bias associated with censorship.
Many models exist for the survival analysis task, including Cox, Aalen, accelerated failure time models, random survival forests, and so on. 
In these models, characterization of proportional errors in rate is often an objective of primary interest, but its minimization is not straightforward because ground truth rates are unobserved.
Indeed, the long-standing success of the Cox proportional hazards model is a demonstration of the interpretive value of proportional rate estimation.
Despite this, the objective function specified for many survival models do not seek to minimize proportional error in rate, including that of Cox.  

We approach the problem of proportional rate minimization for repeated events through the formulation of the objective function.  Our analysis illustrates that the standard likelihood function attends to individuals at highest risk, potentially at the cost of poorly modeling proportional rates in low-risk individuals.  This characterization provides us a way to attempt to mitigate the mis-attention and results in a reweighting scheme to fairly attend to all individuals with respect to proportional rate misspecification.  Practically, this results in a fairness-variance trade-off, as the models suffer from high variance from low effective sample sizes.
We demonstrate in simulations and in prediction of neurological deterioration among patients admitted for intracerebral hemorrhage (ICH) that our method empirically produces informative risk assessments in low rate regimes.

\nocite{miscouridou2018deep,weiss2019recode,avati2018countdown,chapfuwa2018adversarial,jing2017neural,lee2018deephit,yu2011learning}

\section{Method}
\textbf{Background.} 
Let $Y$ be the event we want to model over time across $M$ samples. Let the event times be the sequence $t_{im}$ for $i \in \{1, \dots, T_m\}$ for $m = \{1, \dots, M\}$ over a period of interest $[0, \tau_m]$. We are interested in modeling the rate function:
$$\lambda(t|\cdot) = \lim_{h\rightarrow 0} \frac{P(t<T<t+h|T>t,\cdot)}{h} = \frac{f(t|\cdot)}{S(t|\cdot)},$$
where $\{\cdot\}$ varies by model and represents the information or data to use in modeling $\lambda$. The probability density function and survival function are given by $f$ and $S$. 

Next we establish the relationship between rescaled time where a single Poisson process with rate 1 and our original time, where $\lambda$ is defined.  This comes from the time rescaling theorem.

\textbf{Time rescaling theorem.} Given the rate function $\lambda$, define $\Lambda$ the cumulative hazard function: \mbox{$\Lambda(t|\cdot) = \int_0^{t} \lambda(\tau|\cdot)d\tau$}. For the realization of a sequence of events from $\lambda(t|\cdot)$ with times $\{u_1, \dots, u_k\}$ and $\Lambda(t|\cdot) < \infty$, the sequence $\{\Lambda(u_1|\cdot), \dots, \Lambda(u_k|\cdot)\}$ is distributed according to a unit rate Poisson process \citep{meyer1971demonstration,ogata1981lewis}.

Details of the proof can be found in \citep{brown2002time}. %
The implication of the theorem is that if we could model the conditional intensity correctly, the intervals between rescaled times follow exponential distribution with rate 1. %

\textbf{Harmoic Mean Point Processes.} From the time rescaling theorem, it is straightforward to observe that the relative contributions to the likelihood of each time interval is proportional to the rate within that interval, \emph{i.e.}, if we care about each individual's risk in a time unit equally, then we could consider decreasing the likelihood contributions in proportion to the rate.  In other words, our procedure will seek to nullify, partially or fully, the proportional factor of likelihood attention given to higher rates.   We call this approach optimization of the adjusted log likelihood, which is illustrated in Figure \ref{fig:rescale}.

\begin{equation}
    ALL(\mathbf{X}|\Theta) = \sum_{m=1}^M \Big( \sum_{i=1}^{T_m} \frac{\log \lambda(t_{im}|\Theta)}{\lambda^{*}(t_{im})} - \int_0^{\tau_m} \frac{\lambda(t|\Theta)}{\lambda^{*}(t)} dt \Big)
\end{equation}

where $\lambda^{*}(t)$ is the ground truth intensity at time $t$. By assuming $\lambda(t)$ and $\lambda^{*}(t)$ are piecewise constant, we can view the adjusted log-likelihood as the weighted sum of log-likelihood contributions. Suppose we divide the time interval $(0,\tau_j]$ into $K$ sub-intervals where $K$ is a significantly large number so that $\lambda^{*}(t)$ is constant within any sub-interval. That is, with $0={t'}_{j,0} < {t'}_{j,1} < \cdots < {t'}_{j,K} = \tau_j$, $\lambda^{*}(t)$ is constant for $t\in ({t'}_{j,k-1},{t'}_{j,k}]$ and all $k=1,\ldots,K$. Then,
\begin{align}
    ALL(\mathbf{X}|\Theta)
    &= \sum_{j=1}^M \left[ \int_0^{\tau_j} \frac{\log \lambda(t|\Theta)}{\lambda^{*}(t)}dN(t) - \int_0^{\tau_j} \frac{\lambda(t|\Theta)}{\lambda^{*}(t)} dt \right]\\
    &= \sum_{j=1}^M \sum_{k=1}^K \Bigg[ \int_{{t'}_{j,k-1}}^{{t'}_{j,k}} \frac{\log \lambda(t|\Theta)}{\lambda^{*}(t)}dN(t)
    - \int_{{t'}_{j,k-1}}^{{t'}_{j,k}} \frac{\lambda(t|\Theta)}{\lambda^{*}(t)} dt \Bigg]\\
    &= \sum_{j=1}^M \sum_{k=1}^K \frac{1}{\lambda^{*}({t'}_{j,k})}\Bigg[ \int_{{t'}_{j,k-1}}^{{t'}_{j,k}} \log \lambda(t|\Theta) dN(t)
    - \int_{{t'}_{j,k-1}}^{{t'}_{j,k}} \lambda(t|\Theta) dt \Bigg] \label{eq:weighted_sum}
\end{align}
is the sum of log-likelihood contributions in time intervals $({t'}_{j,k-1},{t'}_{j,k}]$ weighted by the reciprocal of the ground truth intensity at ${t'}_{j,k}$ for every $j$ and $k$. The similarity is apparent when it is compared to the similar form of standard log-likelihood:
\begin{align}
    LL(\mathbf{X}|\Theta) 
    &= \sum_{j=1}^M \sum_{k=1}^K \left[ \int_{{t'}_{j,k-1}}^{{t'}_{j,k}} \log \lambda(t|\Theta) dN(t) - \int_{{t'}_{j,k-1}}^{{t'}_{j,k}} \lambda(t|\Theta) dt \right]
\end{align}
Therefore, we can weight each interval's log likelihood by the inverse of the oracle rate to get the adjusted log likelihood. 

\ConceptPseudocode

\textbf{Oracle Approximation.} Without access to $\lambda^{*}$, however, we must resort to approximation of the reweighting.  One choice for $\lambda^{*}$ is our current estimate $\hat{\lambda}$. However, this could lead to unstable weightings because a single example could dominate the weight distribution.  To address this fairness-variance tradeoff, we introduce the attention coefficient $\gamma$ and stability factor $\epsilon$ to help stabilize the weights.  Pseudocode in Figure \ref{fig:rescale} (right) illustrates the training procedure and the stabilization modification.  We call our method harmonic mean point processes (HMPP) because if the oracle is known and doubly stochastic (frail), then the estimates we get from the training procedure are harmonic mean estimates of the rate distribution.  Note that in practice when we use an approximation, the denominator must be copied and detached from the computation graph so that the graph of which $\lambda$ is a part is not further connected by the current model's predictions $\hat{\lambda}$.

\section{Experimental Setup and Results}
We test our method in two simulations, where we have access to ground truth rates, and in application to a health setting, where we illustrate important factors and effects of our approach.  In all cases, we are comparing our objective versus the standard variant, and call the models Harmonic Mean Point Processes (HMPP, ours) and Maximum Likelihood Point Processes (MLPP, comparison).  We use this labeling across multiple models and domains which we describe next.

\textbf{Simulations.} To test our ability to accurately determine the rates of low risk individuals, we developed a singly- and a doubly- stochastic univariate model with rates varying by 4-6 orders of magnitude ($10^4$ to $10^6$ fold variation in rates). In each case, events are sampled for 10 units of time according to a sample-specific fixed rate $\lambda^{*}$, where $\lambda^{*}$ is drawn from a truncated, base 10 exponential between $10^{-2}$ to $10^{2}$. For the singly-stochastic model, we sample events according to an exponential with rate $\lambda^{*}$, and for the doubly-stochastic model, we sample events according to an exponential with rate $\lambda^{*}10^u$ for $u \sim \text{Uniform}(-2, 0)$.  Then, a time-stamped sequence is produced, containing tuples of \mbox{(id, time, event, value)} features, with the first tuple containing \mbox{(id, 0, rate, $\lambda^{*}$)}.
We use an embedding LSTM architecture for the simulations \citep{dong2018musegan} and provide details in the Appendix.

\textbf{Application: obtundation in intracerebral hemorrhage.} 
We apply our method to real data of neurological decline during intracerebral hemorrhage (ICH) critical care admissions. 
Intracerebral hemorrhage is a life-threatening extravasation of blood outside the vessel wall due to a tear or rupture that results in a blood accumulation, which then presses upon brain tissue causing neuronal damage.  Mortality rates are 40\% at 1 month post diagnosis.  In these individuals, monitoring of neurological status is essential.  The Glasgow Coma Score (GCS) is a score based on physical exam that assesses progression and recovery.  For ICH, GCS is a primary indicator for mortality stratification, intracranial pressure monitoring, and intubation \citep{parry2013accuracy,manno2012update}.  %

\textbf{Data.} We used data from MIMIC III v1.4 \citep{johnson2016mimic}, an EHR housing critical care data on 40,000 individuals. Of those, 1,010 had a primary diagnosis of ICH and were considered as members of our cohort. Chart, laboratory, medications, vitals, procedures, and demographics tables were extracted as time-varying features for nowcasting GCS decreases. The full outcome specification for obtundation (decreased GCS) is in the Appendix.  Figure \ref{fig:descriptives} depicts the study characteristics and approach. We use wavelet reconstruction networks (WRNs) \citep{weiss2018clinical} with the objective function modification to accommodate the adjusted log likelihood.  
\begin{figure}[t]
\centering
    \begin{minipage}{0.5\textwidth}
        \centering
        \TableDemographics
    \end{minipage}%
    \hfill
    \begin{minipage}{0.47\textwidth}
        \includegraphics[width=\columnwidth]{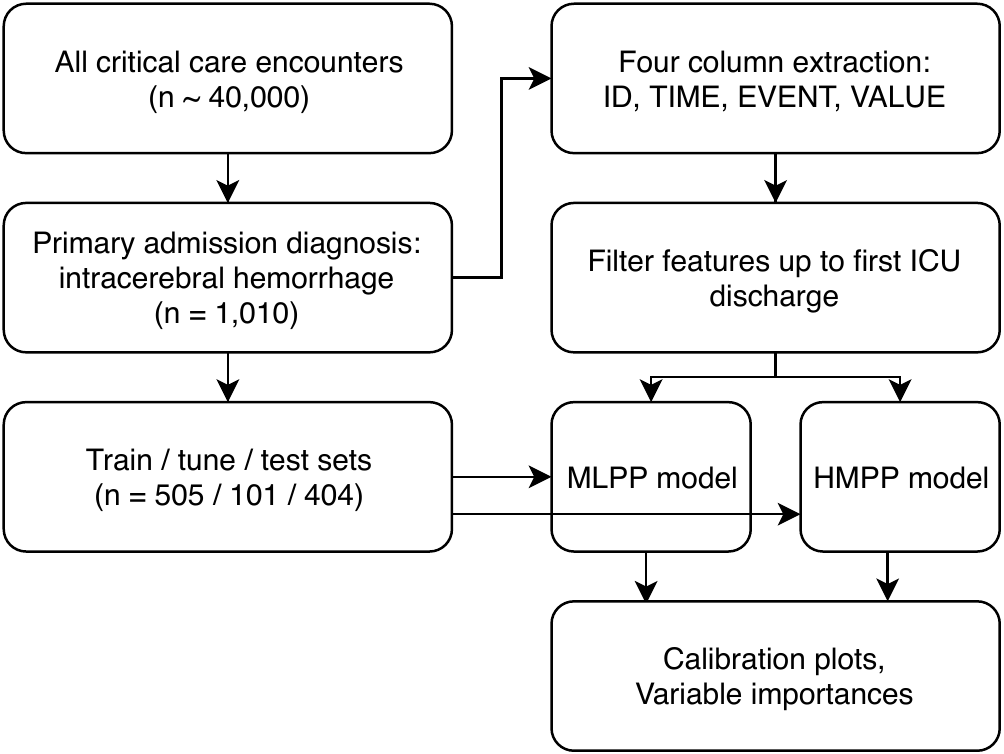}
    \end{minipage}%
\caption{ICH study population characteristic (median, [IQR] or count (\%)) and flowchart.}
\label{fig:descriptives}
\end{figure}
We investigate the performance of the algorithms using inspection of calibration and variable importance plots.

\begin{figure}[t]
\centering
\vspace*{-1.4em}
\includegraphics[width=0.40\columnwidth]{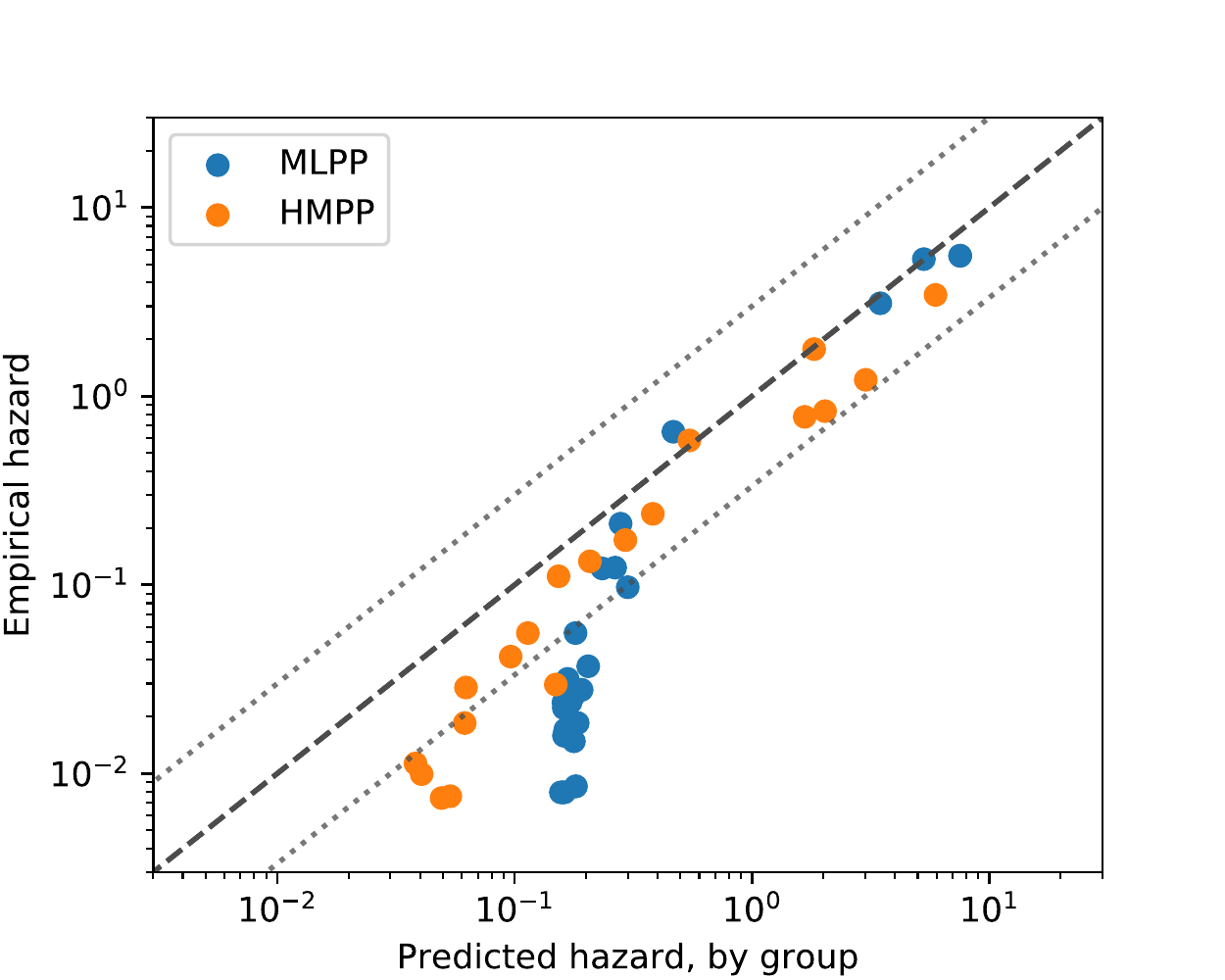}
\includegraphics[width=0.40\columnwidth]{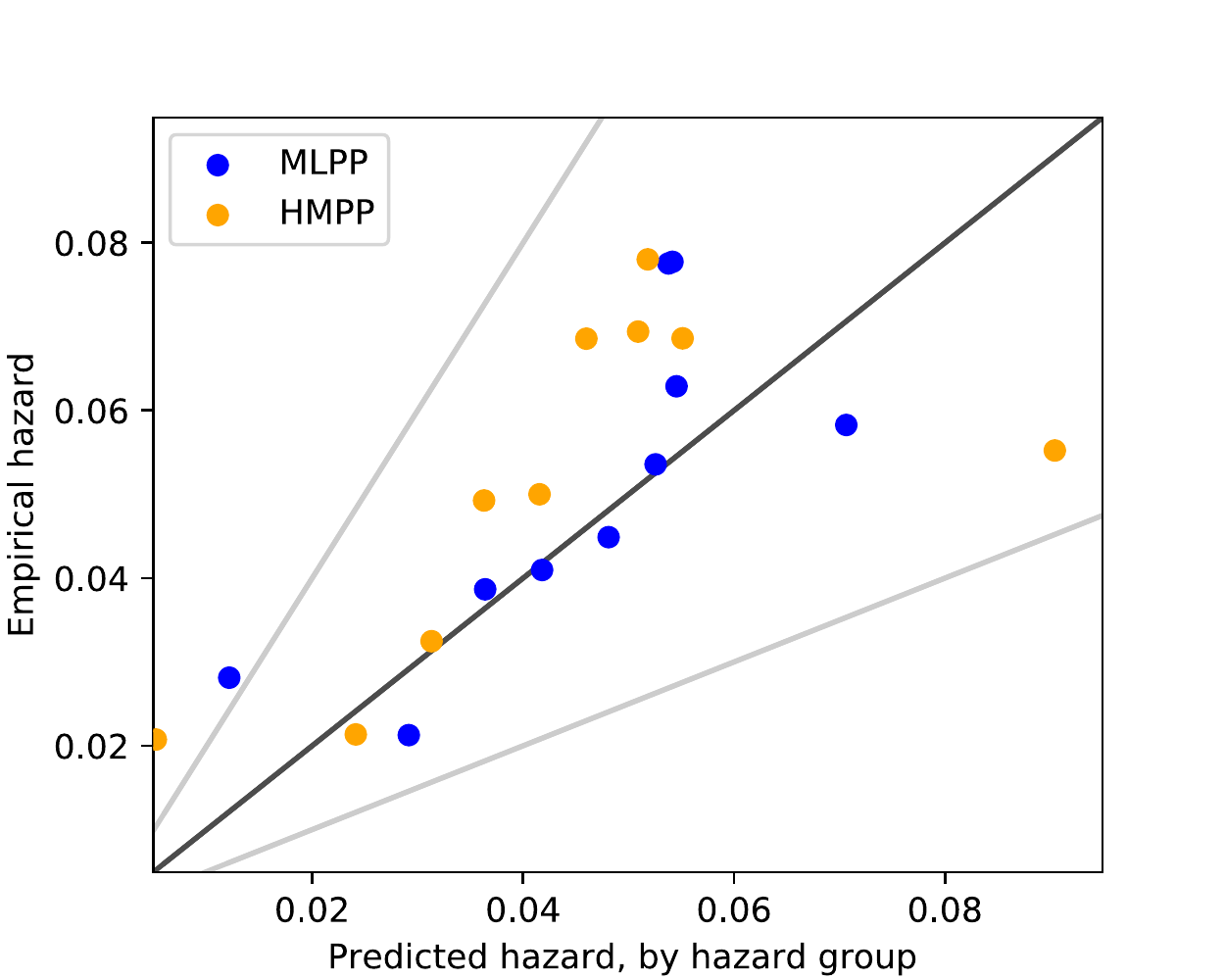}
\caption{Calibration plots comparing HMPP and MLPP in simulation, singly stochastic (left), and in prediction of ICH in MIMIC (right). In simulation, HMPP predicts groups of individuals with rates an order of magnitude smaller. For fixed effects this comes from improved calibration. In MIMIC, lower
rate groups are identified in HMPP, at the cost of higher variability in predictions.}
\label{fig:sim2}
\end{figure}

\textbf{Results.} Figure \ref{fig:sim2} demonstrates the benefit of our method in simulations. In the singly-stochastic model (left figure), both the HMPP and MLPP approaches discriminate risk across the spectrum, illustrated by the (approximately) monotonic curves.  However, the MLPP method never predicts hazards lower than 0.2 for any group, despite many of those groups having empirical rates near 0.01.  By contrast, the HMPP method straightens the low-risk tail and makes more accurate predictions in low risk individuals.  At the same time, risk predictions for high risk individuals are similar in quality. The range of predicted risks from HMPP was half an order of magnitude larger than MLPP.

In the ICH study, the hyperparameters chosen were an elastic net formulation L1 and L2: $10^{-2}$ with $\gamma=10$ and $\epsilon=0.01$, suggesting the model is constrained by limited sample size.
Discriminatively, the C statistic among the lowest quartile was 0.68 (0.62-0.74, bootstrap CI 95\%) and 0.66 (0.60, 0.72) for HMPP and MLPP respectively.
Thus, while HMPP does identify lower risk groups, the small sample size limits the interpretation.  %
Additional results and comments are in the Appendix.

\textbf{Conclusion.}
Our work demonstrates a new tool to make risk predictions in low-risk populations. %
We provide a formulation that exhibits how to attend equally across risk, and provide an algorithm and guidance to trade off fair attention with variance from reweighting.  Importantly, our method detects individuals an order of magnitude lower than predictions made by optimization with the log likelihood and deep network--the combination of two popular approaches.  We further illustrate implications of attending to low-risk individuals in the variable importances reported under each optimization.  This difference may have important applications in suggesting risk factors that are stratum-specific, which can provide guidance in personalized decision making.  %
Future work will include explicit characterization of the proportionate attention-variance tradeoff which could provide alternative approximations to the oracle rate with desirable properties.

\bibliographystyle{apalike}
\bibliography{hmpp.bib}
\newpage
\appendix
\renewcommand\thefigure{\thesection.\arabic{figure}}    
\setcounter{figure}{0}  

\section{Proportional rate error minimization in single failure-time prediction}
There are methods for single failure-time prediction that optimize for the proportional predictive error. For example, accelerated failure time (AFT) models accomplish this by log transforming time so that minimizing mean squared error in log space corresponds to minimizing multiplicative errors in the original space.  While this approach works for single failure-time analyses, it fails to extend to repeated event models.  In particular, AFT models require a specification of time $t=0$ so that the log transformation is well-defined. Time $t=0$ specification is problematic in time-varying and nowcasting analysis, as noted in \citep{miscouridou2018deep} and \citep{weiss2018clinical}, because training time $t=0$ must be specified while the modeler may want to vary test time $t=0$ or model repeated events.
Attempts to duplicate training samples longitudinally by varying train time $t=0$ lead to samples that violate the parametric and or semi-parametric model assumptions and result in poor parameter estimates and poor predictive performance. Because of this problem, it is unclear how to naturally extend such models to recurrent event and multitask settings.  In these settings, larger ranges of rates are often modeled, and this magnifies the problem of misplaced attention.

Another approach is to fuse together binary classification predictions across time.  Our method may help illustrate the effectiveness of this approach, in that this multi-task prediction formulation attends implicitly to low-risk examples by ignoring those examples where events have already occurred in the large time-to-event classifiers \citep{yu2011learning}.  However, unlike these methods, our method provides explicit attention to low-risk persons and regions in the repeated event setting.  

\section{Sensitivity to hyperparameters $\gamma$ and $\epsilon$}
\begin{figure}[b]
\centering
\includegraphics[width=0.49\columnwidth]{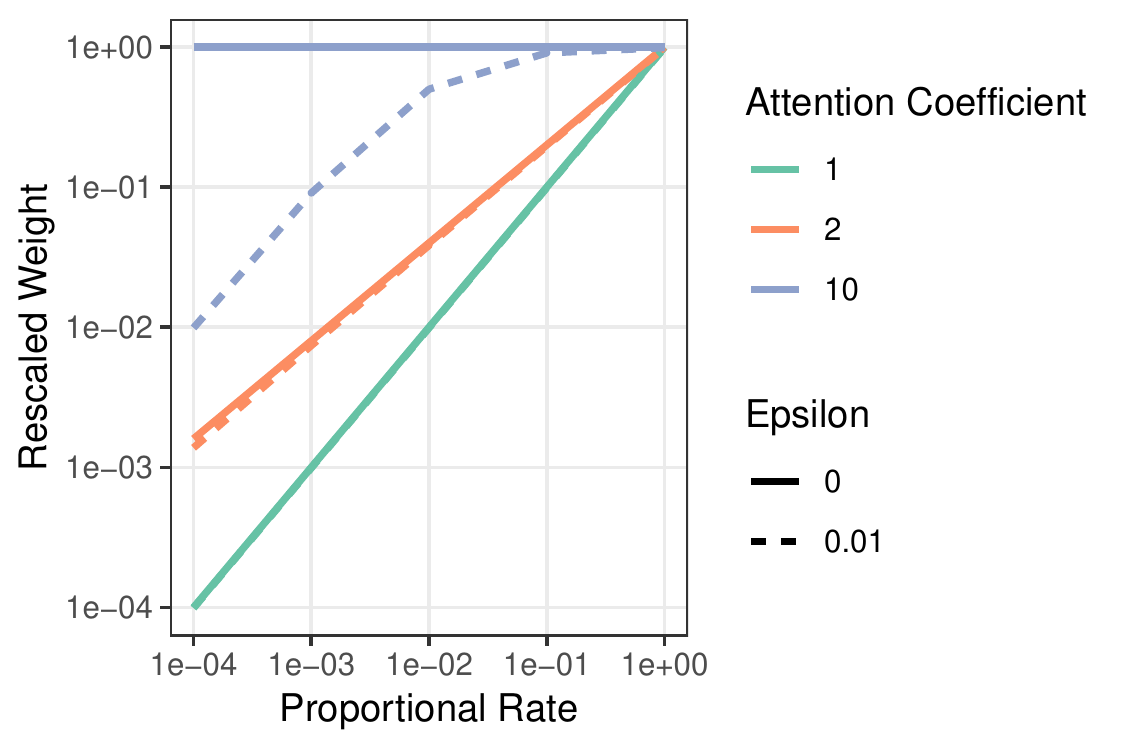}
\includegraphics[width=0.49\columnwidth]{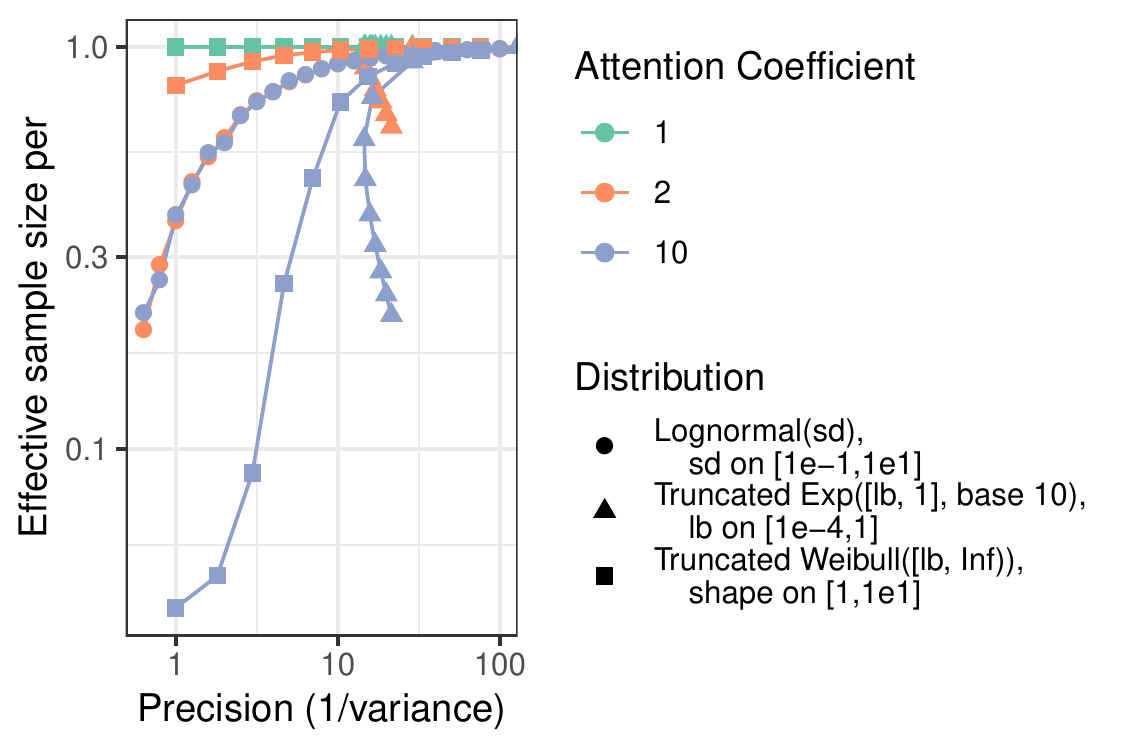}
\caption{Rescaled-space weight based on relative predicted weight, attention coefficient $\gamma$ \mbox{$(\text{MLPP}:\gamma=1)$}, and attention stability factor $\epsilon$ (left). Effective sample size per example under distributional assumption of rates (right).} \label{fig:propweight}
\end{figure}

Figure \ref{fig:propweight} demonstrates the reweighting achieved with different choices of attention coefficient $\gamma$ and stability factor $\epsilon$. To achieve equal rescaled-time weighting, $\gamma$ must be set to 10, corresponding to 10-fold increased weight per 10-fold decrease in risk: the blue horizontal line.  However, the number of effective samples may become very small, shown by the number of effective samples (per 1) for several common distributions.  To avoid this, $\gamma$ and or $\epsilon$ can be chosen to flatten the reweighting distribution.  In practice the predicted distribution is implicit and potentially unstable, and using domain knowledge to set $\epsilon$ near to the lowest rate expected to be found will mitigate the instability while still attending to the low risk individuals.

\section{Architecture} 
\begin{figure}[!ht]
    \centering
    \begin{minipage}{0.62\textwidth}
        \includegraphics[width=\columnwidth]{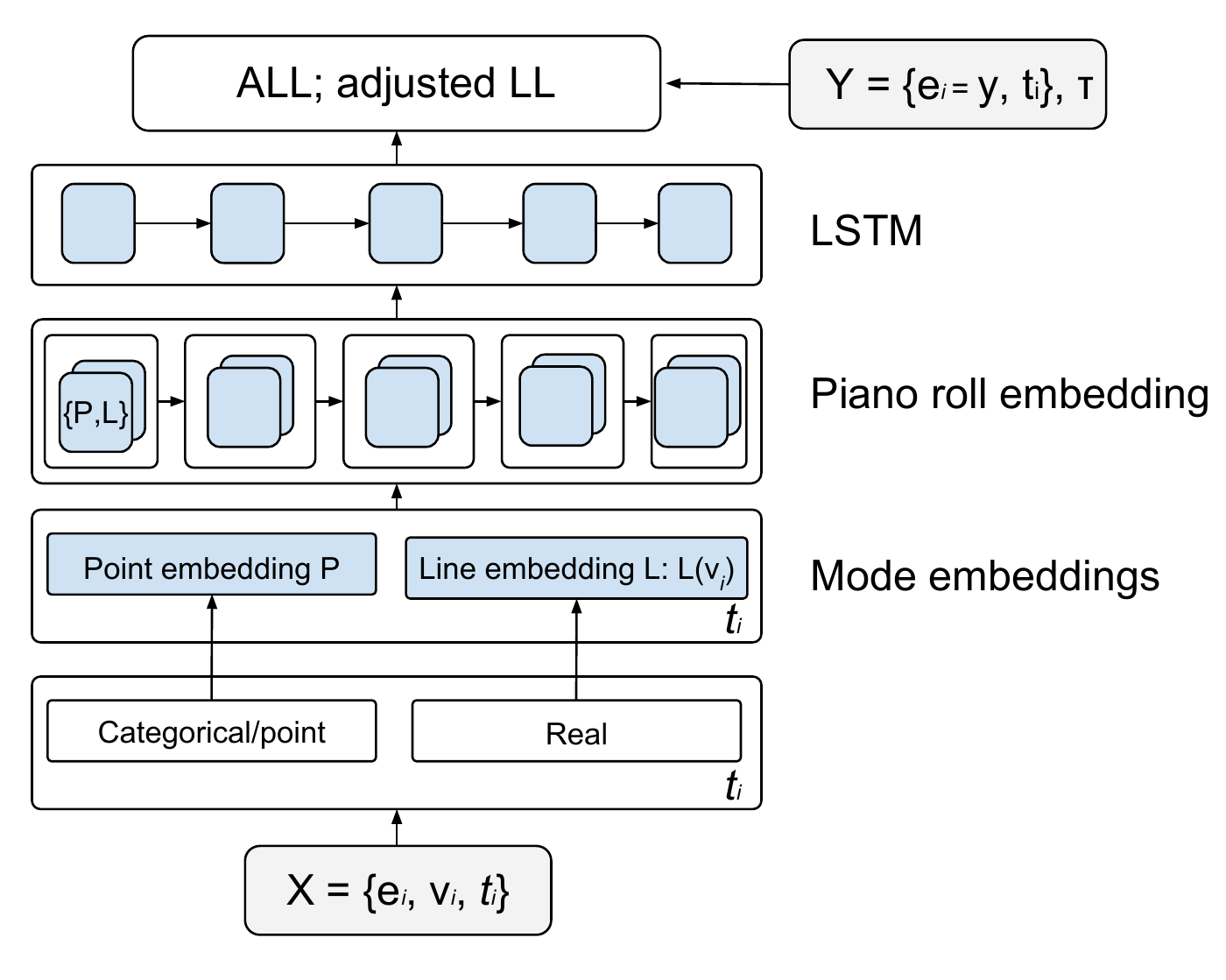}
    \end{minipage}%
\caption{LSTM embedding architecture used for simulations.} \label{prp_architecture}
\end{figure}

 The architecture is given in Figure \ref{prp_architecture}.  The idea is to use a LSTM (piano roll) embedding architecture, where any number of events with or without values can first be embedded as line and point embeddings respectively, and then the embedded signals are captured in a group embedding and passed into LSTM time steps.  This architecture facilitates flexible parsing of long format data typical of marked point processes, such as that of digital orchestral music, or that of medical event streams.  Categorical events are treated as multiple point events, and point events are embedded as points.  Real-valued events are embedded based on their value, and so the event's value domain corresponds to a line embedded as a 1-dimensional manifold. These embedded vectors are then further embedded as a group based on their timestamps, and fed into an LSTM that outputs non-negative rate predictions.  We use times steps of unit length with 10 steps in total.
 
\section{Training hyperparameters}
We conduct training for 50 epochs using the Adam optimizer with a learning rate of $10^{-3}$ and a batch size of 8. The reweighting at each training step also makes across-step ALL comparisons not meaningful.  Therefore, when choosing early stopping points, we use the tune set log likelihood.  We also used the tune set performance in our search over the following hyperparameters: $\gamma \in \{10, 2\}$ (ICH only), $\epsilon \in \{0, 10^{-2}\}$ (ICH only), L1 regularization (LASSO) coefficient in $\{10^{-2}, 10^{-3}, 10^{-4}\}$, and L2 regularization (ridge) coefficient in $\{0, 10^{-2}\}$. Our implementation is in PyTorch v1.0.

\section{Additional methods and results}

\textbf{Outcome definition.}
GCS scores were recorded in the chart table in two versions depending on the vendor, one by the component scores Eyes, Verbal, and Motor, and the other as an aggregate score.  We defined a decrease in GCS to be a decrease of any score not a result of intubation, or a first GCS below 8 (an obtunded state indicative of poor outcomes).  GCS readings inside the critical care unit were considered only.  Per individual, events within the first ICH encounter only were used.

\textbf{Calibration subgroups and variable importance.}
We construct calibration plots using the ordering given by the algorithm predictions, which illustrates discriminative ability in the algorithms ability to stratify groups across the spectrum of risk (an alternative is to order by ground truth rates where the specific expressed goal of assessment is calibration). In applications to real data, we cannot access ground truth, so we order by and use equal quantiles with respect to the predictions.
We also expect important features to vary by objective function and inspect their similarity with variable importance plots.

\begin{figure}[t]
\centering
\includegraphics[width=0.49\columnwidth]{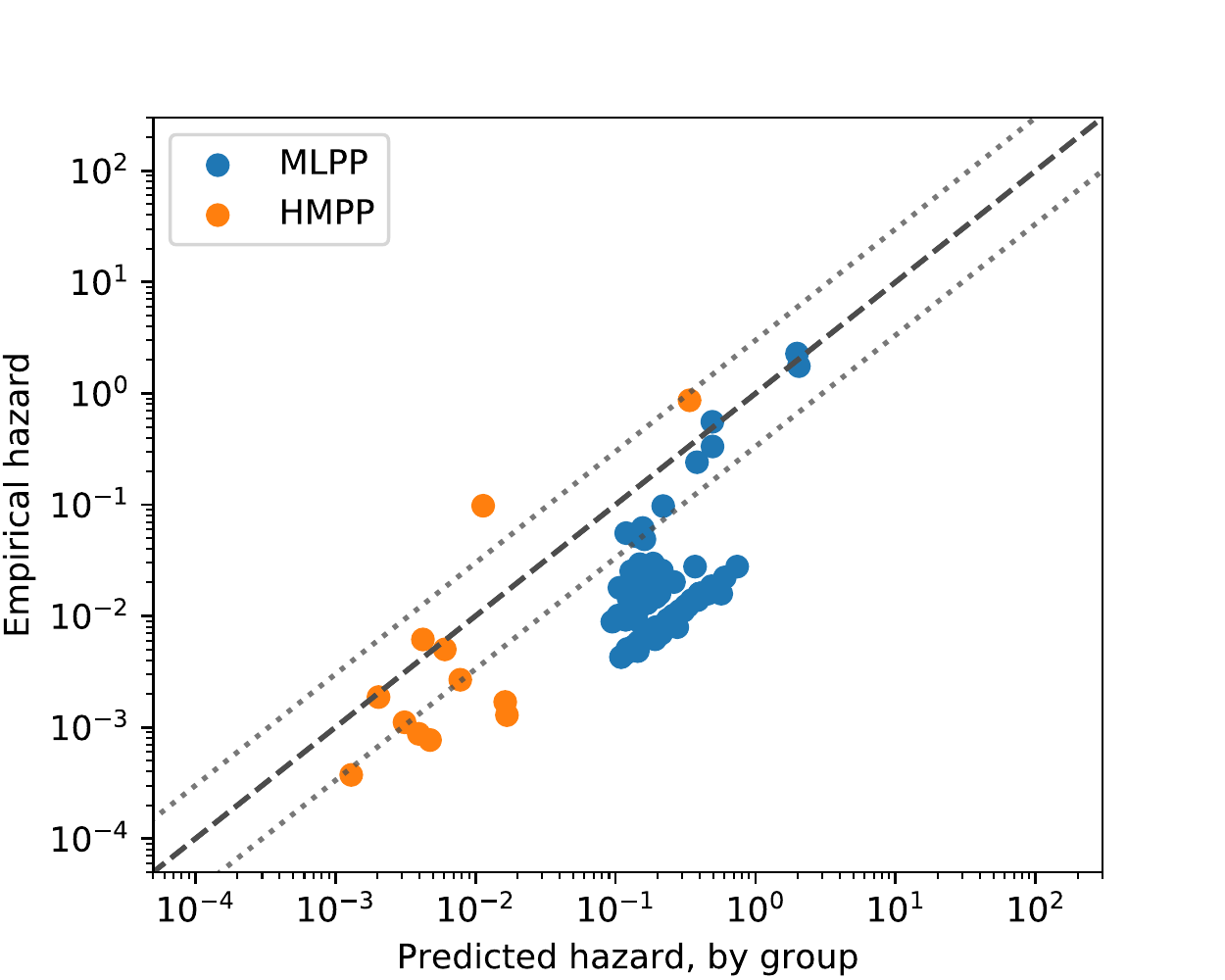}
\caption{Calibration plots comparing HMPP and MLPP in simulation and doubly stochastic processes.  HMPP predicts groups of individuals with rates an order of magnitude smaller. For random effects, this comes from discrimination between the very low- and low- risk individuals. } \label{fig:sim2a}
\end{figure}

\textbf{Additional results.}
For the doubly-stochastic model where the formulation includes frailty, the performance of HMPP and MLPP diverges further. In particular, Figure \ref{fig:sim2a} shows that in the face of random effects that vary the rate ranges by 100-fold, MLPP focuses on the high end of the random effect distribution and HMPP the low end. HMPP identifies groups of individuals with empirical rates an order of magnitude smaller.  It also identifies groups at larger rates, but appears to underestimate the rates for these individuals.  This could be due to overfitting of the training data leading to erroneously low predictions on the test set. Nonetheless, HMPP detects low risk individuals in this setting whereas MLPP does not acknowledge their low rates, instead limiting all predictions to greater than 0.1.

\begin{figure}[t]
\centering
\vspace{1.1em}
\centering
\begin{tabular}{@{\extracolsep{5pt}} lll} 
\hline \\[-1.8ex] 
HMPP & MLPP & MLPP (HMPP par.) \\ 
\hline \\[-1.8ex] 
IV solution & Nasogastric fluid & Foley catheter \\ 
Orogastric fluid & Functional fibrinogen & Osmolality:blood \\ 
Osmolality: blood & Urine white blood cell count & Temperature \\ 
Vancomycin level: blood & Osmolality: blood & Functional fibrinogen \\ 
IV dextrose in water: D5W & IV drip labetalol & IV phenytoin \\ 
Total cholesterol & Vancomycin level: blood & Phenytoin: blood \\ 
Lymphocyte count: csf & Urine red blood cells & IV normal saline \\ 
Urine sodium & Basophil count:blood & IV sterile water added \\ 
IV normal saline & Urine ketones & PO metoprolol \\ 
\hline \\[-1.8ex] 
\end{tabular} 

\caption{Top 10 variable importances for HMPP (left), MLPP (right), and MLPP with HMPP regularization parameter settings (bottom). Note the minimal overlap in the lists, and also note the smaller losses in HMPP, illustrative of earlier stopping during HMPP training.} \label{fig:ich2a}
\end{figure}
Even in the small data setting of ICH in MIMIC, where the low-risk group is not newly identified by the method, we can look at the variable importance plots to demonstrate a marked difference in result. Figure \ref{fig:ich2a} shows HMPP and MLPP variable importances for the top 10 features.  Two features overlap, osmolality and vancomycin levels. For the rest, the HMPP model is concerned with lab tests and intravenous solution choices and quantities while the MLPP model is concerned about urine and clotting lab tests.  It could be that the need for stability in the form of increased regularization and early stopping could be necessary for the HMPP model, so we additionally computed the variable importances for the MLPP model with the same regularization settings, the top ten variables are also shown, which share no increased overlap with the fair model.  This illustrates that the factors that influence proportional risk across the risk spectrum may differ substantially from those obtained from simple likelihood optimization which attends to high risk.

\end{document}